\documentclass[letterpaper]{article} % DO NOT CHANGE THIS
\usepackage{aaai25}  % DO NOT CHANGE THIS
\usepackage{times}  % DO NOT CHANGE THIS
\usepackage{helvet}  % DO NOT CHANGE THIS
\usepackage{courier}  % DO NOT CHANGE THIS
\usepackage[hyphens]{url}  % DO NOT CHANGE THIS
\usepackage{graphicx} % DO NOT CHANGE THIS
\usepackage{natbib}  % DO NOT CHANGE THIS AND DO NOT ADD ANY OPTIONS TO IT
\usepackage{caption} % DO NOT CHANGE THIS AND DO NOT ADD ANY OPTIONS TO IT
\usepackage{algorithm}
\usepackage{algorithmic}

\usepackage{tcolorbox}  % For the tcolorbox environment
\usepackage{fvextra}
\usepackage{newfloat}
\usepackage{listings}
\DeclareCaptionStyle{ruled}{labelfont=normalfont,labelsep=colon,strut=off} % DO NOT CHANGE THIS
\floatstyle{ruled}
\newfloat{listing}{tb}{lst}{}
\floatname{listing}{Listing}
\title{Beyond QA Pairs: Assessing Parameter-Efficient Fine-Tuning \\ for Fact Embedding in LLMs}
\author{
    Shivam Ratnakar\textsuperscript{\rm 1, 2},
    Abhiroop Talasila\textsuperscript{\rm 1},
    Raghav Chamadiya\textsuperscript{\rm 1},\\
    Nikhil Agarwal\textsuperscript{\rm 1},
    Vinayak K Doifode\textsuperscript{\rm 1}
}
\affiliations{
    \textsuperscript{\rm 1}Equinix, 
    \textsuperscript{\rm 2}University of Southern California\\
    shivam.ratnakar@usc.edu
}

\usepackage{bibentry}
\begin{document}

\maketitle

\begin{abstract}
This paper presents an extensive examination of Parameter-Efficient Fine-Tuning (PEFT) for embedding domain specific facts into Large Language Models (LLMs), focusing on improving the fine-tuning process by categorizing question-answer (QA) pairs into `Factual' and `Conceptual' classes using a BERT-based classifier. Two distinct Llama-2 models are fine-tuned based on these classifications and evaluated using larger models like GPT-3.5 Turbo and Gemini. Our results indicate that models trained on conceptual datasets outperform those trained on factual datasets. Additionally, we compare the efficiency of two synthetic fine-tuning dataset generation techniques, D-RAG and D-Naive, with D-Naive demonstrating superior performance. Although PEFT has shown effectiveness, our research indicates that it may not be the most optimal method for embedding facts into LLMs. However, it has demonstrated exceptional performance in instruction-based tasks. Our findings are reinforced by a 1000-sample dataset in the data center domain, where the fine-tuned Llama-2 7B model significantly outperforms the baseline model in generating product recommendations. Our study highlights the importance of QA pair categorization and synthetic dataset generation techniques in enhancing the performance of LLMs in specific domains.
\end{abstract}


\section{Introduction}

\begin{figure*}[h!]
\centering
\includegraphics[width=\textwidth]{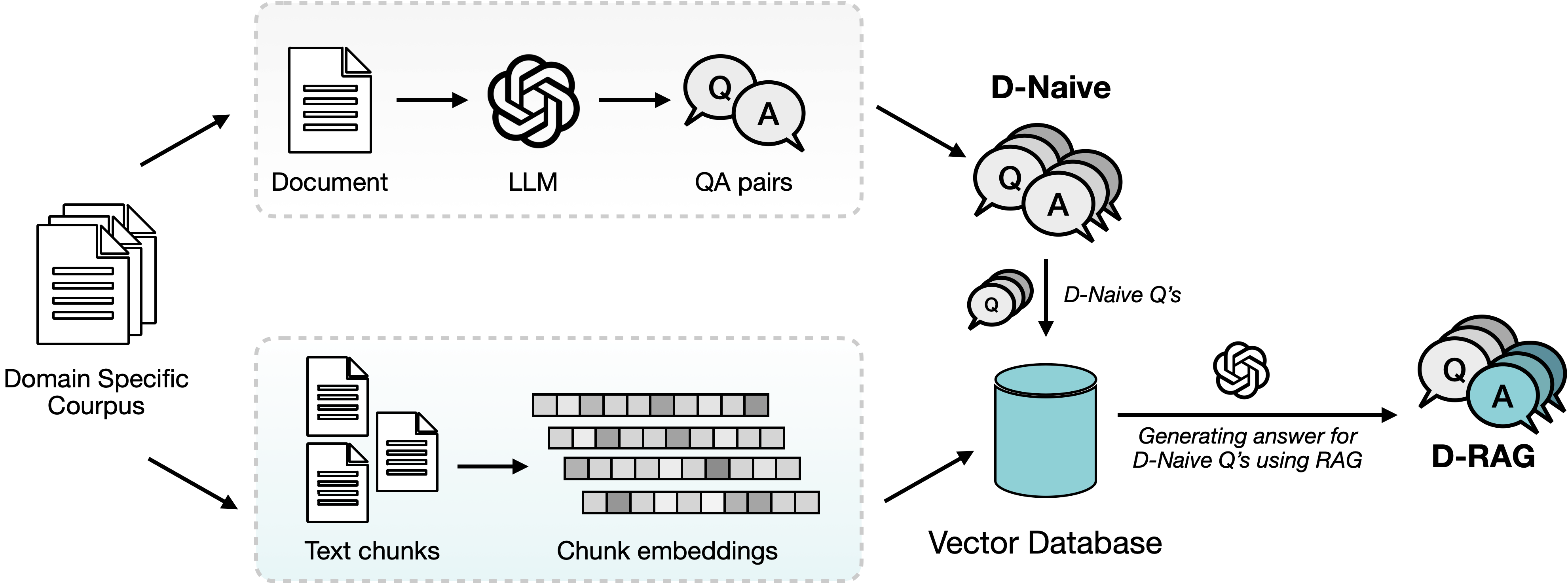}
\caption{Pipeline to generate D-RAG and D-Naive}
\label{fig:drag_dnaive_pipeline}
\end{figure*}

Parameter-Efficient Fine-Tuning (PEFT) has emerged as a highly effective strategy for refining Large Language Models (LLMs) on domain-specific data, thanks to its reduced computational and time requirements compared to full fine-tuning. This technique has seen widespread adoption in the industry for embedding domain knowledge into LLMs. Platforms like Azure, Google Cloud Platform, Mistral, AWS, and Lamini offer fine-tuning as a service using methods like Low Rank Adaptation (LoRA), making PEFT accessible and user-friendly \cite{hu2021lora}. These low code/no code solutions have become popular among developers due to their simplicity.
However, the ease of use of these platforms can create a misconception that merely having a large quantity of question-answer (QA) pairs is sufficient for effective domain adaptation. This misunderstanding may lead to the utilization of low-quality datasets, compromising the effectiveness of the fine-tuning process.
In this paper, we address this issue by proposing a set of metrics to assess the quality and appropriateness of QA datasets for PEFT. We introduce a novel method for categorizing QA pairs into `Factual' and `Conceptual' classes using a BERT-based classifier. By separating the original dataset based on these categories, we fine-tune two distinct sets of Llama-2 models using LoRA. Our evaluation, conducted with larger models such as GPT-3.5 Turbo, Gemini 1.5 Pro \cite{reid2024gemini}, and Prometheus \cite{kim2024prometheus}, reveals that models trained on conceptual datasets significantly outperform those trained on factual datasets.
Furthermore, we investigate the effectiveness of two synthetic dataset generation techniques, D-RAG and D-Naive (depicted in Figure~\ref{fig:drag_dnaive_pipeline}). Our results show that the D-Naive approach produces superior fine-tuning datasets compared to D-RAG. Additionally, we suggest that while PEFT is highly effective, it may not be optimal for embedding factual information into LLMs. Instead, it excels in instruction-based tasks. To support our assertion, we conducted an experiment using a 1000-sample dataset for sales product recommendation in the data center domain. The results clearly demonstrate that the fine-tuned Llama-2 7B model \cite{touvron2023llama} outperforms the baseline model.

\section{Background: Fine-tuning LLMs for Domain Adaptation}

Domain adaptation of LLMs involves tailoring LLMs, initially trained on diverse and extensive public datasets, to enhance their performance and relevance for specific domains or use cases. This process is critical for organizations aiming to leverage LLMs and their reasoning capabilities to address unique concepts and knowledge pertinent to their fields.

There are primarily two approaches to creating domain-specific LLMs: training from scratch or adapting existing general LLMs through continued pre-training. The former is typically cost-prohibitive and less common unless there is a highly specialized requirement. The latter approach, involving fine-tuning existing LLMs, is more efficient and increasingly accessible due to advancements in fine-tuning methodologies and tools.

A significant portion of domain adaptation efforts has concentrated on fine-tuning LLMs using domain-specific QA datasets \cite{zhou2024lima, gupta2024rag, li2023quantity}. These datasets are commonly generated through pipelines where another LLM (e.g., GPT-3.5 Turbo or GPT-4 Turbo) extracts QA pairs from domain resources like text documents, wiki pages, and web content. Alternatively, datasets can be manually curated and annotated by domain experts or compiled from community-driven Q\&A websites like Reddit, WikiHow, and Stack Exchange, often combining multiple approaches for comprehensive coverage. 

 High-quality data selection has therefore garnered significant attention and indicates that only a small fraction of high-quality data may be necessary to achieve performance comparable to fine-tuning on entire datasets. Techniques to optimize data selection have been developed, with algorithms and heuristics identifying the most effective data subsets, enhancing efficiency and reducing resource requirements for domain adaptation \cite{zhou2024lima, chen2023maybe, shen2024rethinking, li2023quantity}.

\section{Is PEFT suitable for embedding facts?}

PEFT techniques like LoRA affect a very small fraction of the weights primarily in the Self-Attention module, whereas LLM's knowledge is thought to be stored in the Feedforward Network module. Therefore, our hypothesis is that these techniques are suitable for influencing the writing style or incorporating specific logic into the text generation process of LLMs. For instance, in domain-specific text summary generation, crafting sales pitches from product descriptions, selecting the most suitable product based on descriptions and customer requirements, etc. These use-cases are better suited for PEFT instead of use-cases where the LLM needs to learn certain facts about the domain. To test this hypothesis, we generate two types of datasets. First category represents the task that would require a business logic to be embedded into the LLM or some change in the LLM's text generating style. The second category of datasets represents factual information based QA bot for a specialized domain. With these datasets, we trained a Llama-2 7B model and compared the performance of on the two categories. We also argue that the change in writing style category of use-cases requires very little amount of data when compared to factual embedding use-cases. Through are experiments, we also showcase the effectiveness of conceptual QA pairs over factual QA pairs for domain specific QA bot use-cases. Our experiments indicate that a question like ``What is a patch panel?'' is a better data point for fine-tuning in comparison to ``How many patch panels are there in the XYZ Silicon Valley data-center?''.

\section{Related Work}

The task of optimizing data quality for fine-tuning LLMs intersects with various facets of model training, including the evaluation of minimal data requirements, domain adaptation, and instruction tuning. This section highlights approaches that have informed the development of our proposed framework.

\paragraph{Minimal Data Requirements and Efficiency} Recent studies emphasize the efficacy of fine-tuning LLMs with minimal but high-quality datasets. The LIMA model \cite{zhou2024lima} demonstrated that fine-tuning a 65B parameter Llama model with just 1,000 curated prompts yielded performance on par with models trained on extensive datasets, underscoring that most knowledge is acquired during pretraining. Similarly, ``Maybe Only 0.5\% of Data is Needed'' \cite{chen2023maybe} explores reducing data usage in instruction tuning, revealing that models can achieve better task-specific performance with significantly less data, challenging the necessity of large datasets for fine-tuning.

\paragraph{Domain-Specific Adaptation} Fine-tuning for domain-specific tasks has been extensively explored. The RAFT approach \cite{zhang2024raft} combines Retrieval-Augmented Generation (RAG) with fine-tuning to enhance LLM performance in specific domains by training models to disregard irrelevant retrieved documents, improving focus and accuracy. Similarly, ``RAG vs Fine-tuning'' \cite{gupta2024rag} compares both approaches across various LLMs, demonstrating how each method can be effectively employed for domain-specific applications, particularly in underexplored sectors like agriculture. Additionally, ``Fine-tuning Language Models for Factuality'' \cite{tian2023fine} leverages recent innovations in factuality judgment and preference optimization algorithms to improve the factual accuracy of LLMs, offering a novel approach to mitigating misinformation.

\paragraph{Instruction Tuning and Data Selection} Efficient data selection for instruction tuning is crucial for optimizing LLM performance. ``From Quantity to Quality'' \cite{li2023quantity} introduces a self-guided methodology that employs the Instruction-Following Difficulty metric to identify high-quality instruction data, enhancing training efficiency. Additionally, ``Rethinking Data Selection for Supervised Fine-Tuning'' \cite{shen2024rethinking} argues that selecting data reflecting human-like interactions, rather than purely based on quality and diversity, yields better results in aligning models with human expectations. The MoDS approach \cite{du2023mods} further refines data selection by focusing on quality, coverage, and necessity, demonstrating improved performance with a significantly reduced dataset. Addressing LLM limitations such as hallucinations and weak numerical reasoning, ToolQA \cite{zhuang2024toolqa} introduces a dataset to evaluate LLMs' ability to use external tools for question answering, providing insights into their strengths and weaknesses.

\begin{table*}
  \centering
  \begin{tabular}{llll}
    \hline
    \textbf{Methods} & \textbf{GPT} & \textbf{Gemini} & \textbf{Prometheus} \\
    \hline
    D-RAG & $3.67 \pm 1.504$ & $2.72 \pm 1.43$ & $3.23 \pm 1.60$ \\
    D-Naive & $\textbf{3.93} \pm \textbf{1.073}$ & $\textbf{2.81} \pm \textbf{1.22}$ & $3.19 \pm 1.40$ \\
    Factual & $3.62 \pm 1.178$  & $2.24 \pm 1.32$  & $2.81 \pm 1.41$ \\
    Conceptual & $\textbf{4.02} \pm \textbf{1.213}$ & $\textbf{2.84} \pm \textbf{1.26}$  & $\textbf{3.34} \pm \textbf{1.33}$ \\
    \hline
  \end{tabular}
  \caption{\label{tab:res_table} Average evaluation scores and standard deviation of different evaluator LLMs on our four QA datasets}
\end{table*}

\section{Experiments and Results}

In order to test the hypothesis of ineffectiveness of PEFT on factual embedding based use-cases like QA bots, we generated 5 datasets that can be used to fine-tune an LLM. All these datasets were generated using GPT-4 Turbo. The data for generating QA pairs and prompt response pairs was scraped from publicly available websites of a data center company. These websites belong to different logical groups, ensuring that the datasets cover a diverse range of information within the specific domain. The authors manually reviewed the generated QA/prompt-response pairs to eliminate any erroneous data points from the datasets. The first four of these five datasets represent the use-case of QA bots and the last one is a product recommendation dataset. We describe these datasets as follows:

\begin{itemize}
\item \textbf{D-RAG and D-Naive} Figure~\ref{fig:drag_dnaive_pipeline} visualizes our synthetic dataset creation technique, depicting the pipeline for generating two QA datasets, D-Naive and D-RAG, from a domain-specific corpus. In the D-Naive approach, an LLM is used to directly generate QA pairs from documents. The process is straightforward, wherein each document is fed into the LLM prompted to generate QA pairs. On the other hand, the D-RAG approach uses RAG over the questions generated by the D-Naive method and regenerates answers by querying the vector database. This approach enhances the quality of answers by leveraging the entire corpus's context rather than relying on single documents. This essentially improves the QA pairs generated by D-Naive by providing more contextually rich answers. Each of these datasets contain 20,000 QA pairs, out of which 1000 pairs were used for testing.
\item \textbf{Conceptual and Factual} These datasets were derived from the D-Naive dataset, with each containing 5,000 QA pairs. The total dataset consists of 10,000 QA pairs, which is a subset of the original 20,000 QA pairs in the D-Naive dataset. The dataset is divided into two classes: conceptual and factual. The conceptual dataset consists of questions that require a deeper understanding of the domain rather than relying solely on factual knowledge. An example of a conceptual question is ``What is a patch panel?'' On the other hand, the factual dataset contains questions that necessitate specific factual knowledge about the domain. An example of a factual question is ``How many XYZ Inc. data centers are located in California?'' A BERT-based classifier was used to classify the pairs as conceptual or factual. This classifier was trained on a corpus of 5,000 QA pairs annotated by GPT-4 Turbo. Training the BERT-based classifier helped save time and cost in annotating QA pairs at scale.
\item \textbf{Product Recommendation} This dataset consists of 1000 prompt-response pairs created from product descriptions scraped from the websites. The prompts include a list of products and a data-center requirement, and the LLM is tasked with selecting the most suitable product for the given requirement.
\end{itemize}

\paragraph{Fine-tuning} A Llama-2 7B model from \texttt{NousResearch} hosted on HuggingFace was fine-tuned on each of these datasets using LoRA. Training parameters were optimized to balance performance and learning efficiency. Notable configurations included a training and evaluation batch size of 8 per device, gradient accumulation across four steps, and gradient check pointing to enhance memory efficiency. The training included 5 epochs with an initial learning rate of 2e-4, employing mixed precision $(bfloat16)$ to expedite computation. AdamW with blockwise model-update filtering was used as the optimizer, and a cosine scheduler managed the learning rate with a warm-up ratio of 5\%. 

We opted for the Llama-2 7B model due to its widespread availability and compatibility with low-code and no-code fine-tuning platforms. This particular model has gained significant traction among industry developers, who frequently utilize it to construct domain-specific question-and-answer bots. Consequently, we sought to assess the effectiveness of PEFT on this model for these specialized applications. By doing so, we aim to understand its potential and performance in real-world scenarios, thereby contributing valuable insights to the ongoing discourse in the field. 

\paragraph{Proctor LLMs} Proprietary LLMs like GPT-4 Turbo are widely used to assess the quality of responses from various LLMs. However, issues with transparency, tuning, and cost highlight the need for open-source LLMs specialized in evaluation. Current proctor LLMs often produce scores that diverge significantly from human ratings and are limited to general criteria like helpfulness and harmlessness, lacking the ability to assess based on custom evaluation metrics.

We use three proctor LLMs: GPT-3.5 Turbo, Gemini 1.5 Pro \cite{reid2024gemini}, and Prometheus 2 7B \cite{kim2024prometheus} to score our generated QA datasets according to a predefined rubric (Evaluation Prompt in Appendix). Prometheus has demonstrated strong evaluation capabilities, closely aligning with both human and GPT-4 Turbo assessments while being significantly smaller in size. 

Our evaluation framework systematically compares the performance of these proctor LLMs. Each model scored a diverse set of QA pairs generated by various LLMs, using a consistent rubric that included custom evaluation metrics.

\paragraph{Evaluation results} The mean and standard deviation of LLM scores of all 4 QA pairs datasets are shown in Table~\ref{tab:res_table}. The proctor models compare the output generated by the LLM against the ground truth answer in the test dataset. The test dataset consists of 1000 QA pairs each for every category of the training dataset. The prompt and rubric provided to the LLMs to generate these scores are provided in the Appendix.

The findings reveal that the model trained on conceptual data exhibits superior performance compared to the model trained on factual data. Surprisingly, the D-Naive evaluation scores surpass those of D-RAG, contrary to our initial expectation. Further investigation of the D-RAG dataset revealed that the retriever failed to retrieve the appropriate documents, resulting in lower-quality answers compared to D-Naive. In the product recommendation dataset, the trained model successfully recommended the correct product in 70 out of 100 test queries. In contrast, the vanilla Llama-2 7B model performed significantly worse, accurately recommending the right product in only 30 queries.

\section{Conclusions}

Our research highlights the paramount importance of the quality and categorization of QA pairs in PEFT, providing profound insights into optimizing the fine-tuning process of LLMs for domain-specific applications. The outcomes of our fine-tuning experiments reveal that PEFT is particularly advantageous for scenarios requiring minimal factual information embedding into LLMs. Notably, the LLM trained on a conceptual dataset significantly outperformed the one trained on a factual dataset. This trend was consistently observed across all three proctor models, underscoring that the sheer volume of QA pairs is insufficient for the effective deployment of PEFT in developing domain-specific QA bots. It is crucial to judiciously select the use-case when leveraging PEFT. Our product recommendation experiment further illustrates that for instruction-based applications, even a dataset as modest as 1,000 prompt-response pairs can yield a high-quality fine-tuned model.

Although our experiments with D-RAG and D-Naive did not demonstrate that the D-RAG technique for synthetic training data generation is more efficient, we believe that this avenue warrants further exploration. The potential of D-RAG to generate more comprehensive and complete answers remains promising. In this particular instance, the technique's shortcomings were primarily due to the suboptimal performance of the vector database retriever. By addressing these retrieval inefficiencies, future research could unlock the full potential of D-RAG, thereby contributing to more effective and nuanced fine-tuning methodologies for LLMs. Thus, while current findings emphasize the importance of careful use-case selection and QA pair quality in PEFT, they also open the door for continued innovation in synthetic data generation techniques.

\section{Limitations and Future Work}
In this paper, our research has been constrained to a knowledge base derived from a single domain. While the findings provide valuable insights into the impact of data quality on PEFT, expanding these experiments to encompass a broader range of domains would significantly enhance our understanding. Such expansion could reveal domain-specific nuances and broaden the applicability of our conclusions.

The techniques and experiments presented herein are inherently versatile and can be applied across various domains. We posit that, irrespective of the domain, instruction-based datasets are inherently more suitable for PEFT. This hypothesis is grounded in our findings, which consistently demonstrated superior performance with conceptual datasets over factual ones.

Looking forward, we plan to incorporate alternative fine-tuning techniques, such as full parameter fine-tuning, particularly for use-cases that require substantial factual information embedding. This comparative analysis will help delineate the strengths and limitations of PEFT relative to other fine-tuning methodologies, providing a more comprehensive framework for optimizing LLMs for diverse applications.

Our current research exclusively employs Llama-2 7B, chosen for its widespread adoption in industry applications, robust performance across various benchmarks, and established credibility. However, to generalize our findings and explore the scalability of our approach, future work will extend these experiments to include other LLMs of comparable parameter sizes, as well as larger models. Evaluating the performance of these models on factual embedding use-cases will provide deeper insights and potentially uncover new avenues for enhancing fine-tuning processes.

By addressing these limitations and pursuing these future directions, we aim to contribute to the ongoing evolution of fine-tuning methodologies, ensuring that LLMs can be more effectively tailored to meet the specific needs of diverse and complex domain-specific applications.

\bibliography{custom}

\appendix
\section{Appendix}
\label{sec:appendix}

\subsection{Training Details}

The experiments were conducted on a high-capacity \texttt{Azure Standard\_NC96ads\_A100\_v4} compute instance, featuring 880 GB of RAM, 4 NVIDIA A100 PCIe GPUs each with 80 GB of memory, for a total GPU memory of 320 GB, and 96 processor cores. This setup was chosen to effectively manage the computational demands of fine-tuning LLMs. 

% The fine-tuning process was executed using the \texttt{NousResearch/Llama-2-7b-hf} model hosted on Huggingface. Training parameters were optimized to balance performance and learning efficiency. Notable configurations included a training and evaluation batch size of 8 per device, gradient accumulation across four steps, and gradient checkpointing to enhance memory efficiency. The training included 5 epochs with an initial learning rate of 2e-4, employing mixed precision $(bfloat16)$ to expedite computation. AdamW with blockwise model-update filtering was used as the optimizer, and a cosine scheduler managed the learning rate with a warm-up ratio of 5\%.

\begin{figure}[t]
  \includegraphics[width=0.9\columnwidth]{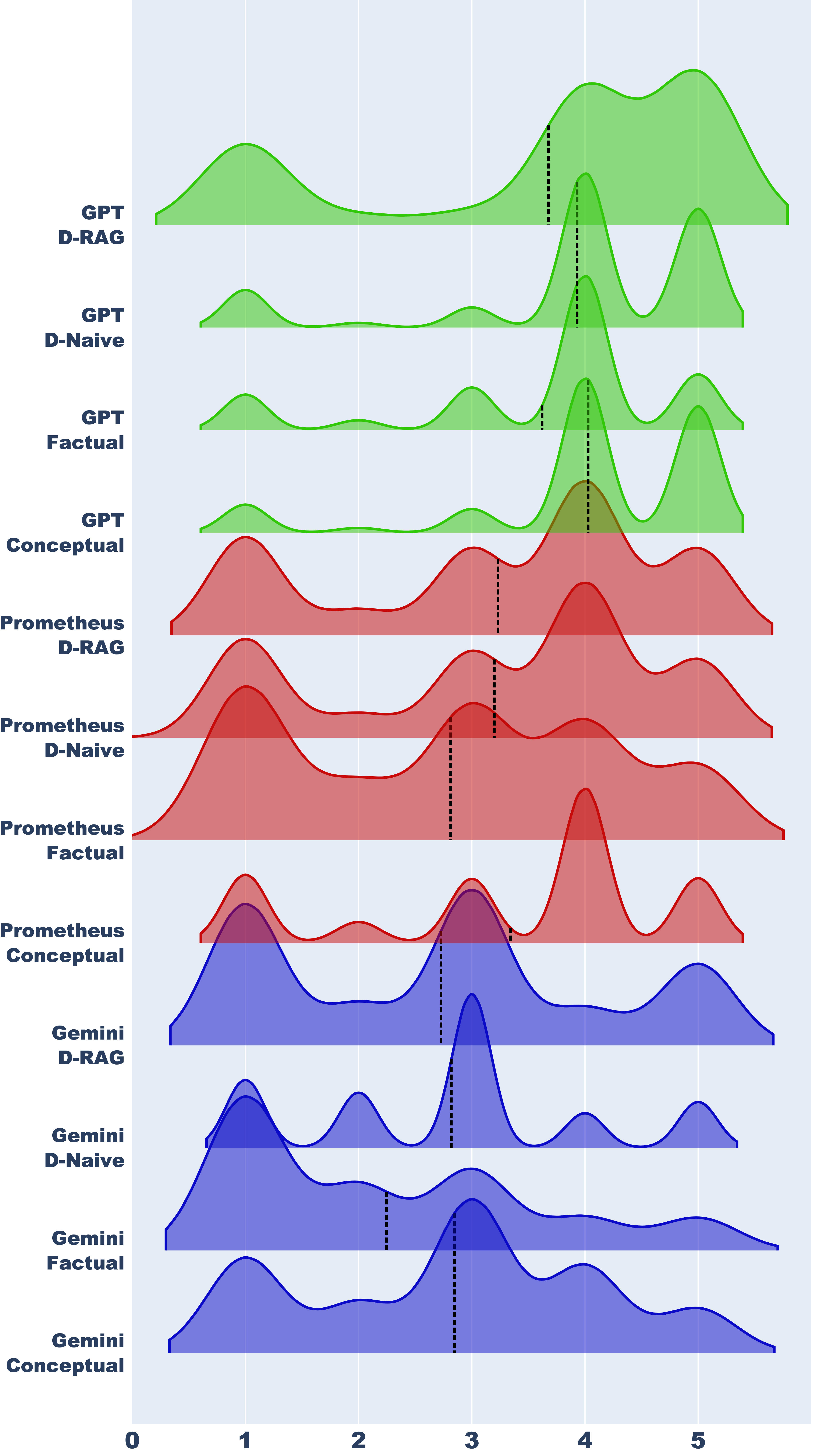}
  \caption{Comparison of score distribution of different evaluators. Refer to Table~\ref{tab:res_table} for empirical results}
  \label{fig:experiments2} 
\end{figure}

% big figures
\begin{figure*}[h!]
\centering
\includegraphics[width=\textwidth]{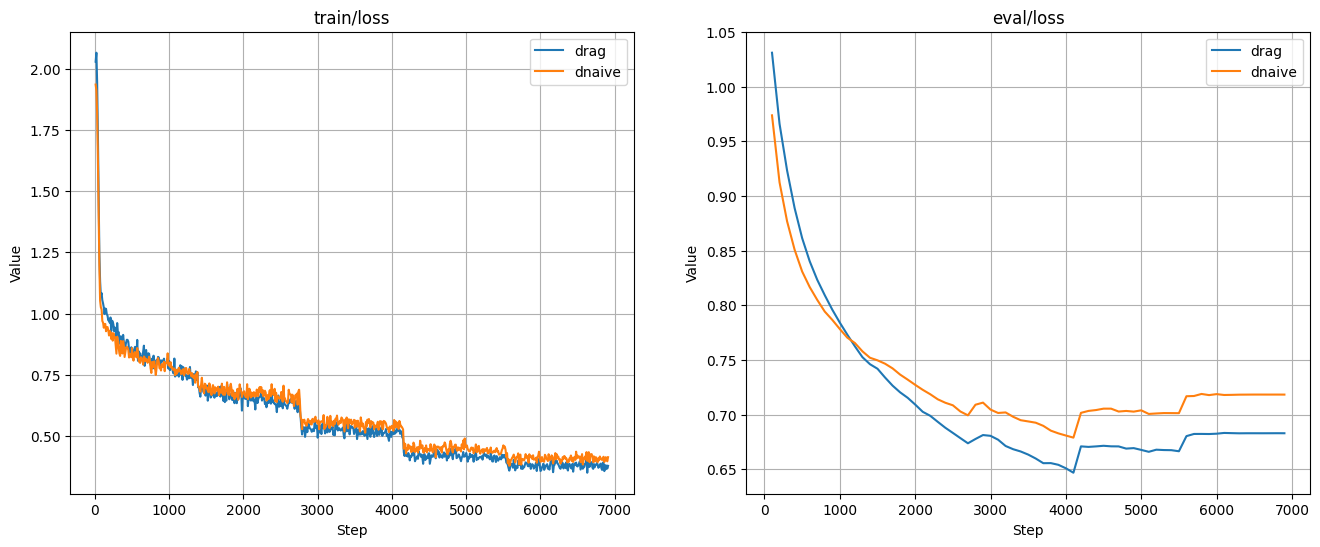}
\caption{Train and Eval loss - D-RAG vs D-Naive}
\label{fig:train-eval1}
\end{figure*}

\begin{figure*}[h!]
\centering
\includegraphics[width=\textwidth]{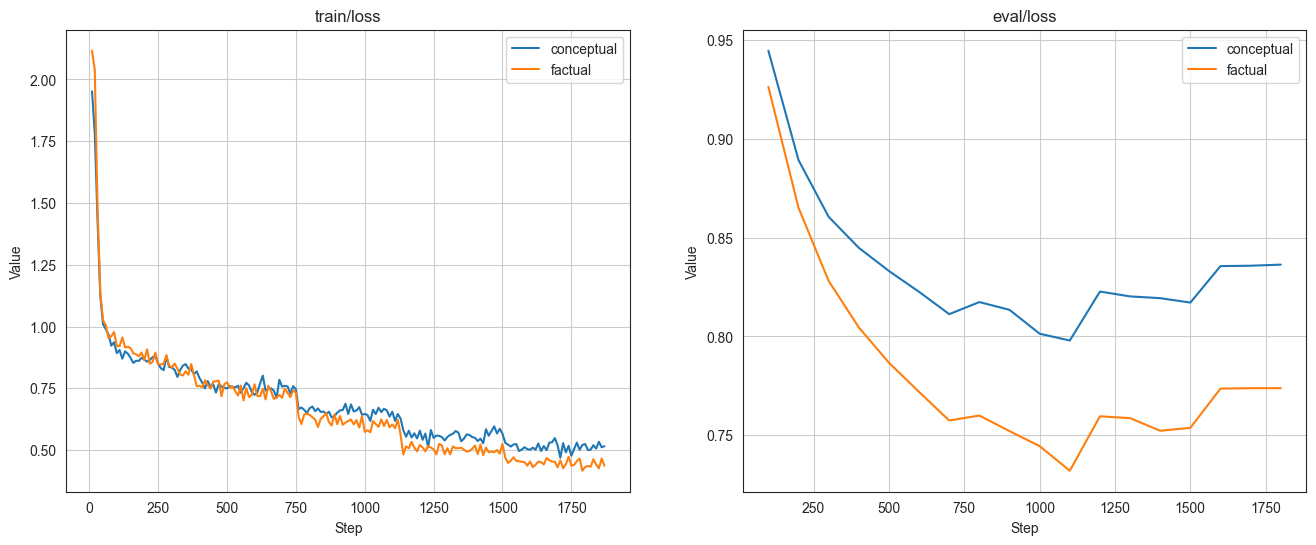}
\caption{Train and Eval loss - Conceptual vs Factual}
\label{fig:train-eval2}
\end{figure*}

\begin{figure}[t]
    \includegraphics[width=\columnwidth]{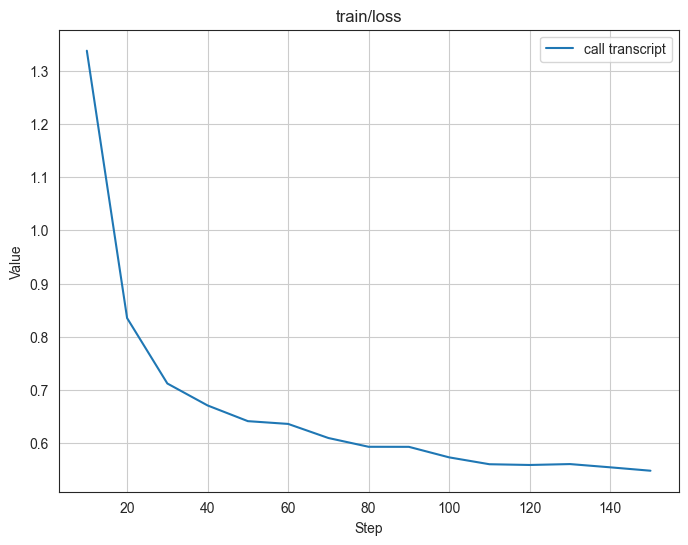}
    \caption{Train loss - Call Transcript}
    \label{fig:train-eval-call}
\end{figure}

\begin{figure}[t]
    \includegraphics[width=\columnwidth]{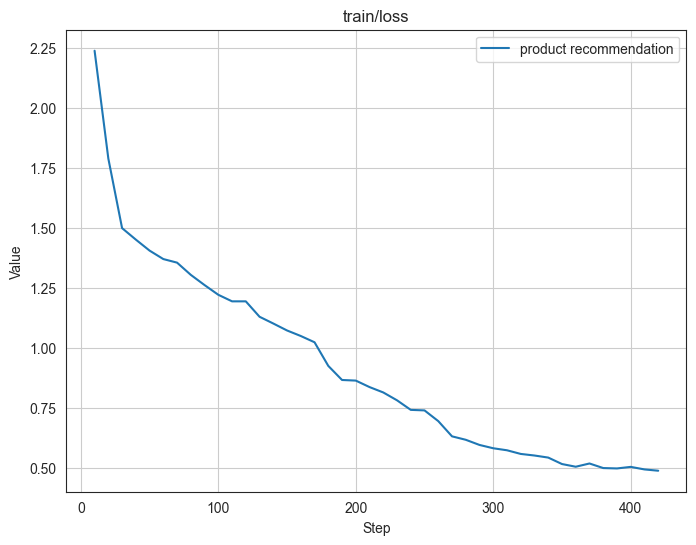}
    \caption{Train loss - Product Recommendation}
    \label{fig:train-eval-product}
\end{figure}

\begin{figure}[t]
    \includegraphics[width=\columnwidth]{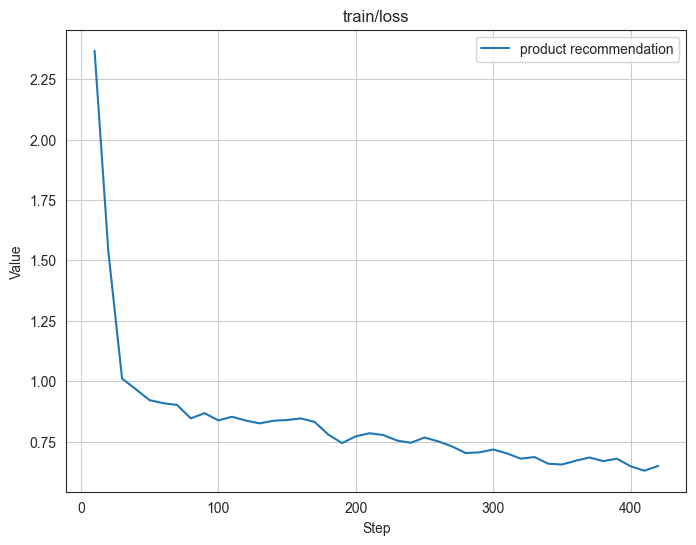}
    \caption{Train loss - Sales Pitch}
    \label{fig:train-eval-sales}
\end{figure}

% Evaluation prompt
%%%%%
\onecolumn
\tcbset{colback=gray!10, colframe=gray!50, width=\columnwidth, arc=2mm, auto outer arc, boxrule=0.5mm}
\begin{tcolorbox}[title=Evaluator LLM Prompt]
\begin{Verbatim}[breaklines=true, breaksymbol={}]
Task Description:
An instruction (might include an Input inside it), a response to evaluate, a reference answer that gets a score of 5, and a score rubric representing a evaluation criteria are given.
1. Write a detailed feedback that assess the quality of the response strictly based on the given score rubric, not evaluating in general.
2. After writing a feedback, write a score that is an integer between 1 and 5. You should refer to the score rubric.
3. The output format should look as follows: Feedback: {{write a feedback for criteria}} [RESULT] {{an integer number between 1 and 5}}
4. Please do not generate any other opening, closing, and explanations. Be sure to include [RESULT] in your output.
 
The instruction to evaluate: {instruction}
 
Response to evaluate: {response}
 
Reference Answer (Score 5): {reference_answer}
 
Score Rubrics:
Is the response correct, accurate, and factual based on the reference answer?
Score 1: The response is completely incorrect, inaccurate, and/or not factual.
Score 2: The response is mostly incorrect, inaccurate, and/or not factual.
Score 3: The response is somewhat correct, accurate, and/or factual.
Score 4: The response is mostly correct, accurate, and/or factual.
Score 5: The response is completely correct, accurate, and factual.
 
Feedback:
\end{Verbatim}
\end{tcolorbox}
%%%%%

% DRAG DNAIVE EXAMPLE
%%%%%
\onecolumn
\tcbset{colback=gray!10, colframe=gray!50, width=\columnwidth, arc=2mm, auto outer arc, boxrule=0.5mm}
\begin{tcolorbox}[title=D-RAG and D-Naive Example]
\begin{Verbatim}[breaklines=true, breaksymbol={}]
Question: How does <COMPANY> Fabric simplify the configuration of networking for a Metal-hosted cluster?

D-RAG Answer: <COMPANY> Fabric simplifies the configuration of networking for a Metal-hosted cluster by providing a global software-defined interconnection platform. It allows you to connect the Metal site to remote cloud or other infrastructure easily. With <COMPANY> Fabric, you can set up a fabric-based networking configuration, which not only makes the setup process easy but also allows you to keep your inter-cluster traffic private if desired. Additionally, <COMPANY> Fabric supports a container network interface (CNI) plugin designed for multi-cluster networking, making it relatively easy to configure. This simplifies the networking setup for Metal-hosted clusters and provides flexibility for future changes to the Kubernetes environment or architecture.

D-Naive Answer: <COMPANY> Fabric simplifies the configuration of networking for a Metal-hosted cluster by providing a global software-defined interconnection platform. By using <COMPANY> Fabric, you can easily connect the Metal site to other remote cloud or infrastructure where your other cluster is hosted. This allows for seamless communication between clusters and eliminates the complexities of configuring networking tools that may vary from site to site. Additionally, <COMPANY> Fabric enables you to keep your inter-cluster traffic private if desired.
\end{Verbatim}
\end{tcolorbox}
%%%%%

% CALL TRANSCRIPT AND NEXT STEP EXAMPLE
%%%%%
\onecolumn
\tcbset{colback=gray!10, colframe=gray!50, width=\columnwidth, arc=2mm, auto outer arc, boxrule=0.5mm}
\begin{tcolorbox}[title=Call Transcript and Next Step Example]
\begin{Verbatim}[breaklines=true, breaksymbol={}]
Sales: Hi, John here from the sales team. I wanted to discuss a new feature idea for our product.
Developer: Hi John, sure. What's the feature you have in mind?

Sales: I was thinking of adding a chatbot feature that can assist customers with their inquiries.
Developer: That sounds interesting. How do you envision the chatbot to work?

Sales: The chatbot should be able to answer frequently asked questions, provide product information, and guide customers through the purchasing process.
Developer: Okay. Are there any specific platforms or technologies you would like us to use for building the chatbot?

Sales: We're open to suggestions, but it would be great if the chatbot can be integrated into our existing website and mobile app.
Developer: Got it. I'll look into the available chatbot frameworks and APIs that we can leverage for this project.

Sales: Perfect. Let's also consider personalization options and the ability to hand over to a human agent if needed.
Developer: Definitely, we can add those features as well. I'll start working on the requirements and come up with a development plan.

Sales: Great! Once we have the chatbot ready, we can demonstrate it to potential customers and gather feedback.
Developer: That's a good plan. We'll need to involve the marketing team to create awareness and promote the new feature.

Sales: Absolutely. I'll reach out to the marketing team and coordinate with them to align our strategies.

Next Steps:
Sales: Coordinate with the marketing team to align strategies and create awareness about the new chatbot feature.
Developer: Research chatbot frameworks and APIs, and come up with a development plan for the chatbot feature.
\end{Verbatim}
\end{tcolorbox}
%%%%%

% PRODUCT DESCRIPTION AND PITCH
%%%%%
\onecolumn
\tcbset{colback=gray!10, colframe=gray!50, width=\columnwidth, arc=2mm, auto outer arc, boxrule=0.5mm}
\begin{tcolorbox}[title=Product description and Pitch]
\begin{Verbatim}[breaklines=true, breaksymbol={}]
Product Description:
<COMPANY> Metal is a global platform that allows users to easily provision bare metal servers anywhere in the world. It provides state-of-the-art hardware without any virtualization layer, offering pure, unadulterated iron. With <COMPANY> Metal, users can experience the ease and convenience of deploying servers from the comfort of their laptop.

Product Pitch:
Experience the power of <COMPANY> Metal, the global platform for bare metal server provisioning. With state-of-the-art hardware and easy deployment process, <COMPANY> Metal allows you to spin up servers anywhere in the world. Say goodbye to virtualization layers and enjoy the pure performance of unadulterated iron.
\end{Verbatim}
\end{tcolorbox}
%%%%%

% PRODUCT RECOMMENDATION
%%%%%
\onecolumn
\tcbset{colback=gray!10, colframe=gray!50, width=\columnwidth, arc=2mm, auto outer arc, boxrule=0.5mm}
\begin{tcolorbox}[title=Product Recommendation]
\begin{Verbatim}[breaklines=true, breaksymbol={}]
For the following customer requirement and given list of product description, output the name of the product which can be recommended to customer to solve their problem

Customer requirement:
The customer requires a networking solution that enables seamless communication between clusters in a multi-cluster Kubernetes environment, running at different sites.

Products:
<COMPANY> Metal offers bare metal servers that are ready to use when and where needed. They provide the flexibility and reliability required to navigate challenging supply chains and ensure on-time delivery of IT hardware.

Network Edge is a service offered by <COMPANY> that allows users to create virtual devices with primary and secondary redundancy. It provides a flexible and scalable networking solution for businesses.

The Unified Cross Connects Portal is a platform provided by <COMPANY> that allows customers to manage and order Cross Connects with ease. It provides a centralized interface for accessing Cross Connect services, enabling customers to efficiently configure and schedule their connections.

<COMPANY> Fabric is a global software-defined interconnection platform provided by <COMPANY>. It allows for easy networking configuration in a multi-cluster Kubernetes environment, enabling seamless communication between clusters running at different sites. With <COMPANY> Fabric, you can ensure secure and private inter-cluster traffic while simplifying your networking setup.

<COMPANY> Metal is a cloud infrastructure offering that allows users to manage Metal resources in event-driven configurations. It provides a growing Ansible collection as a provider for seamless integration.

Output: <COMPANY> Fabric
\end{Verbatim}
\end{tcolorbox}
%%%%%

\end{document}